# Sentiment and position-taking analysis of parliamentary debates: a systematic literature review

**Gavin Abercrombie[1]** · **Riza Batista-Navarro[1]**

## Abstract

Parliamentary and legislative debate transcripts provide access to information concerning the opinions, positions, and policy preferences of elected politicians. They attract attention from researchers from a wide variety of backgrounds, from political and social sciences to computer science. As a result, the problem of computational sentiment and position-taking analysis has been tackled from different perspectives, using varying approaches and methods, and with relatively little collaboration or cross-pollination of ideas. The existing research is scattered across publications from various fields and venues. In this article, we present the results of a systematic literature review of 61 studies, all of which address the automatic analysis of the sentiment and opinions expressed, and the positions taken by speakers in parliamentary (and other legislative) debates. In this review, we discuss the existing research with regard to the aims and objectives of the researchers who work in this area, the automatic analysis tasks which they undertake, and the approaches and methods which they use. We conclude by summarizing their findings, discussing the challenges of applying computational analysis to parliamentary debates, and suggesting possible avenues for further research.

## Introduction

Debate transcripts from legislatures such as the United Kingdom (UK) and European Union (EU) parliaments and the United States (US) Congress, among others, provide access to a wealth of information concerning the opinions and attitudes of politicians and their parties towards arguably the most important topics facing societies and their

✉ Gavin Abercrombie
    gavin.abercrombie@manchester.ac.uk

1   University of Manchester, Manchester, UK

🖄 Springer

citizens, as well as potential insights into the democratic processes that take place in the world's legislative assemblies.

In recent years, these debates have attracted the attention of researchers from diverse fields and research backgrounds. These include, on one hand, computer scientists working in the field of natural language processing (NLP), who have investigated the application and adaptation to the political sphere of methods developed for sentiment analysis of product reviews and blogs, and who have also tackled other related tasks in this domain, such as topic detection. In addition, political and social scientists, traditionally relying on expert coding for the analysis of such transcripts, have increasingly been exploring the idea of viewing 'text as data' [23], and using computational methods to investigate the positions taken by debate participants.

As a result, a wide range of approaches to the problem of automatic debate analysis have been adopted, with research on this problem varying widely in its aims and methods. Within this body of work, there exist many inconsistencies in the use of terminology, with studies in some cases referring to very similar tasks by different names; while in others, the same term may mean quite different things. For example, while Chen et al. [16] and Kapočiūtė-Dzikienė and Krupavičius [32] both attempt to classify debate speakers according to party affiliation, the former refer to this as 'political ideology detection', and the latter as 'party group prediction'. Conversely, a single term like 'sentiment analysis' may be used to refer to, among other things, support/opposition detection [74], a form of opinion-topic modeling [51], and psychological analysis [25]. The approaches adopted range from statistical analyses to predictive methods, including both supervised classification and unsupervised topic modeling. There are also contrasting approaches to modeling the textual data, the level of granularity of the analyses, and, for both supervised learning methods and the evaluation of other approaches, the acquisition and application of labels used to represent the ground-truth speaker sentiment.

With regard to synthesis of the existing research on this topic, Kaal et al. [31] assembled researchers from diverse fields to investigate the problem of text analysis in political texts, Glavaš et al. [22] presented a tutorial addressing similar themes to this review, and both Hopkins and King [26] and Monroe et al. [47] discussed the general differences in the aims and objectives of social scientists and computer scientists when working on such problems. However, as far as we are aware, there exists no comprehensive written overview, systematic or otherwise, of research in this area. The aim, therefore, of this review is to bring together work from different research backgrounds, locating and appraising literature concerning computational sentiment and position-taking analysis that has been undertaken to date on the domain of parliamentary and legislative debate transcripts. We assess the research objectives, the types of task undertaken, and the approaches taken to this problem by scholars in different fields, and present suggested directions for future work in this area.

## Terminology

In the NLP literature, the terms *opinion mining* and *sentiment analysis* are used more or less interchangeably (for a discussion of this, see Pang and Lee [54]), and are employed to describe both the specific task of determining a document's sentiment polarity (that is, *positive* or *negative*, or sometimes *neutral*), and the more general problem area of automatically identifying a range of emotional and attitudinal 'private states' (that is, non-observable, subjective states),[1] such as 'opinion, sentiment, evaluation, appraisal, attitude, and emotion' [39]. In a recent survey, Yadollahi et al. [77] list nine such different sentiment analysis sub-tasks.

Researchers in the political and social sciences, meanwhile, appear to lack a single term to describe the act of determining from text the positions taken by legislators. In the literature, such tasks are variously referred to as 'political scaling', 'position scaling', 'ideal point estimation', and a range of other task- and data set-specific terms. One phrase that appears throughout such work is 'text as data' [e.g., 23, 36, 56]. We, therefore, include this term in our systematic search to capture work from the field that utilizes computational methods for speaker position analysis.

For the purposes of this article, we use 'sentiment analysis' as a general umbrella term, encompassing any tasks concerned with the extraction of information relating to speakers' opinions and expressed positions, and 'sentiment polarity classification' for the more specific, binary or ternary classification task.

### Research questions and objectives

In carrying out this review, we aim to answer the following research questions to ascertain the current state of research in this area:

- RQ1: What are the research backgrounds of the authors of papers published in this area, and to what extent is the work multi-disciplinary?
- RQ2: From which parliaments and other legislatures have debates been analyzed?
- RQ3: What are the objectives of researchers from different backgrounds working on sentiment and position analysis of parliamentary debates?
- RQ4: What sentiment analysis task sub-types have been undertaken to conduct analysis of parliamentary debates?
- RQ5: What approaches have been taken to sentiment/position analysis of parliamentary debates?
- RQ6: What conclusions can be made about the reported performance and outcomes of the sentiment/position analysis systems that have been described?

### Review scope and method

For this review, we followed the established systematic review guidelines of the PRISMA (Preferred Reporting Items for Systematic Reviews and Meta-Analyses) statement [46]. The use of systematic review methodology to conduct this review has enabled us to uncover a substantial body of relevant work, but has meant the exclusion of some potentially interesting studies. We were unable to include a number of known relevant results uncovered by an initial scoping search of the Google Scholar platform, which, due to the lack of transparency of its search algorithm, does not facilitate replication. While somewhat limiting in this sense, the decision to adhere to a systematic methodology provides a replicable and transparent method of synthesizing and summarizing the literature and identifying future research priorities.

---

[1] See Quirk et al. [59].

Although there exists relevant work on related domains such as political campaign speeches [e.g., 43, 49, 71] and electoral manifestos [e.g., 42], we limited our search to publications concerning the automatic analysis of the sentiment, opinions, and positions expressed by participants in the transcripts of debates in parliaments and other legislatures, and also excluded any studies that do not report the results of empirical experiments.

The review covers all literature retrieved by systematic search of five digital library databases and repositories [see Fig. 1(1)]. These were selected as they provided high coverage of the results obtained by the prior scoping search. That search also provided the basis of our keyword search terms, which we developed to return results that included, as a minimum, all the relevant publications previously found (Fig. 1(2)).[2]

All searches were conducted on January 31st, 2019. Following deduplication, screening, and eligibility assessment, 61 studies have been included in the review. Using the NVivo qualitative data analysis software package [63], we coded these according to (a) their research backgrounds, (b) the legislature and language of the debate transcripts analysed, (c) their stated research objectives, (d) the sentiment and position analysis tasks undertaken, (e) the approaches taken and methods used, and (f) the reported performances of the described sentiment/position analysis systems (see Fig. 2). The full-review protocol pipeline is shown in Fig. 1.

## Research backgrounds

We categorize the research background of each study according to the institutional affiliation(s) of its author(s) and the nature of its venue of publication, coding them as either *computer science*, *political/social science*, or *multi-disciplinary*. We consider a study to be multi-disciplinary if it (a) is written by authors from two or more research backgrounds, or (b) the paper is published at a venue associated with a different research background than that of its author(s)' affiliations(s). While it is of course possible that the work which we class as being from computer science or the social sciences actually involves some level of inter-disciplinary collaboration that does not fit within our definition (for example, we do not investigate the authors' academic histories), this is a straightforward yet systematic way of obtaining a general overview of the research community working in this area.

We find that over half the studies are written from a computer science background ($n = 35$). Within this are researchers working on two kinds on problems. First, there are those who approach the transcripts from a computational linguistics perspective,

[2] To reduce the number of results returned, we also added exclusion terms such as 'big data' and 'twitter', which we found did not prevent the retrieval of any known relevant publications.

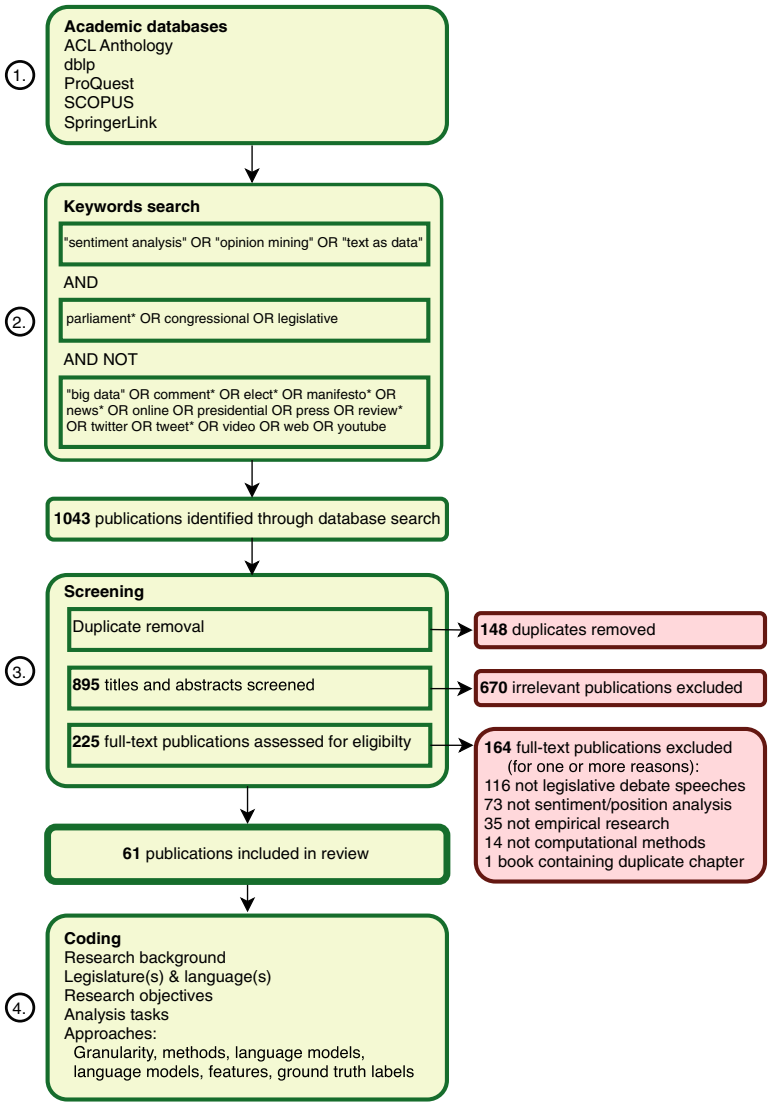

**Fig. 1** Flow diagram of the phases of the systematic review process: 1. database selection; 2. keyword search; 3. screening and eligibility assessment; 4. manual coding

and whose work relates to properties of the language used such as argumentation structures and dialog [18, 48]. The second, larger group consists of work that can be characterized as belonging to the field of NLP, and whose work is more focused on tools and applications [e.g., 30], less than half the number of included studies as computer scientists (*n* = 14), and just 12 studies involve multi-disciplinary research. Of these, seven involve both computer scientists and political or social scientists [32, 35, 61, 62, 64, 65, 75], three collaboration between linguists and computer scientists [25, 28, 51], and two that include researchers from three different fields [17, 50]. According to the number of studies published on this subject annually, interest in this area has been increasing over time, particularly in recent years (see Fig. 3).

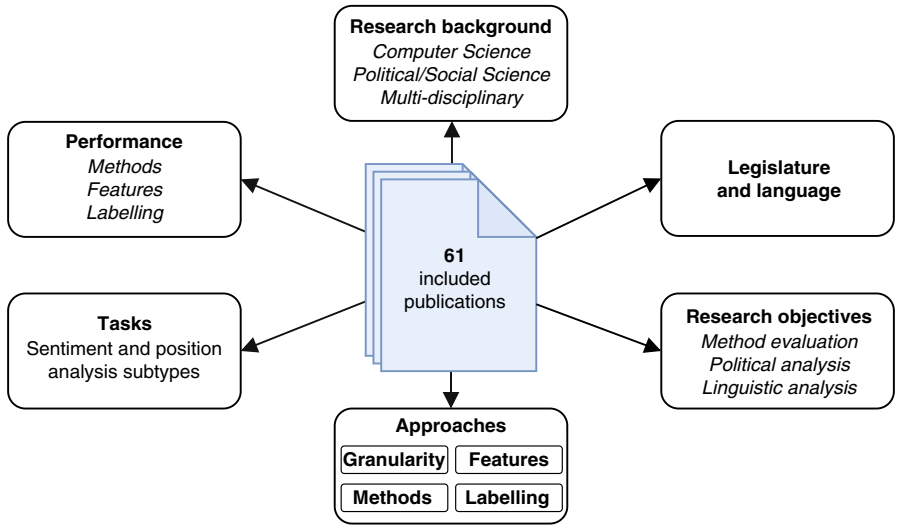

**Fig. 2** Framework for coding the included studies

## Parliaments and legislatures

Nearly, all the included studies focus on one single legislature for analysis, with only Sakamoto and Takikawa [65] and Proksch et al. [56] comparing their approaches (to the analysis of the level of polarization, i.e., ideological division, in parliaments) on transcripts from two or more different chambers. The US Congress is by far the most popular legislature for analysis, attracting the attention of 31 of the studies. This can partly be attributed to the global power and influence of the US and of the English language, but is also explained by the widespread use by NLP researchers of the *ConVote* corpus [74] as a benchmark data set for the evaluation of sentiment analysis systems. Indeed, including its original authors, 17 of the publications use this data set, 15 of which are written from a computer science background, with Hopkins and King [26] (social science) and Iyyer et al. [28] (multi-disciplinary) the exceptions. In some cases, *ConVote* is used alongside one or more other non-legislative data sets (such as product reviews) for the evaluation of text classification methods [5, 15, 16, 28, 30, 38, 41, 79–81]. In fact, only a little over half (37) of the studies are exclusively concerned with the analysis of legislative debates. On the whole, political and social scientists seemingly prefer to construct their own data sets from the congressional record to suit their research aims, while Sakamoto and Takikawa [65] (multi-disciplinary computational social science) also do so.

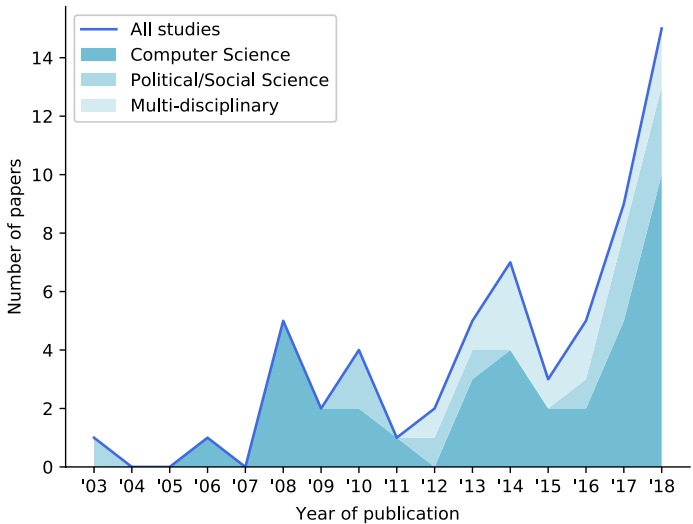

**Fig. 3** Rate of publication of papers featured in the review from 2003, the year of publication of the first paper, to 2018. The review additionally features one paper published in 2019

Following Congress, the next most analyzed legislatures are the UK Parliament (*n* = 9) [1, 2, 18, 52, 62, 66–68, 76] and the EU Parliament (*n* = 5) [20, 21, 25, 56, 57]. The French [4, 37, 55] and Canadian parliaments [3, 48, 61] and the Irish Dáil [36, 40, 56] all appear in three studies, while the California State Legislature [12, 33] and the German Bundestag [56, 60] are both analyzed in two papers. The Austrian [64], Dutch [75], Lithuanian [32], Norwegian [35], Czech, Finnish, and Spanish [56] parliaments, Swiss Federal Assembly [69] Polish Sejm [19], Japanese Diet [65], and the UN General Assembly [8] are all utilized in only one study each (see Fig. 4).

It is notable that, thus far, research in this area appears to have been restricted to data from North America, Europe, and Japan. Sources of transcript data reported in the included studies are shown in Table 1.

Nearly all the included publications consist of analysis of parliamentary or legislative data in a single language. Exceptions include Sakamoto and Takikawa [65] who use two corpora in different languages (English and Japanese), and Glavaš et al. [21] who use a multilingual dataset (German, French, English, Italian, and Spanish), as do Proksch and Slapin [57] (English, French, and German translations). In the latter case, while the original data are multilingual (23 official languages of the European Parliament), the transcripts have been translated to these three languages. Similarly, for speeches not originally in English, Baturo et al. [8] use official translations from the other official languages of the UN (Arabic, Chinese, French, Russian, and Spanish). By far, the most prominent language is English, analyzed in 55 studies. This is followed by French and German, used in six studies each, and a long tail of languages that only appear in one study each (Czech, Dutch, Finnish, Italian, Japanese, Lithuanian, Norwegian, Polish, Spanish, and Swedish).

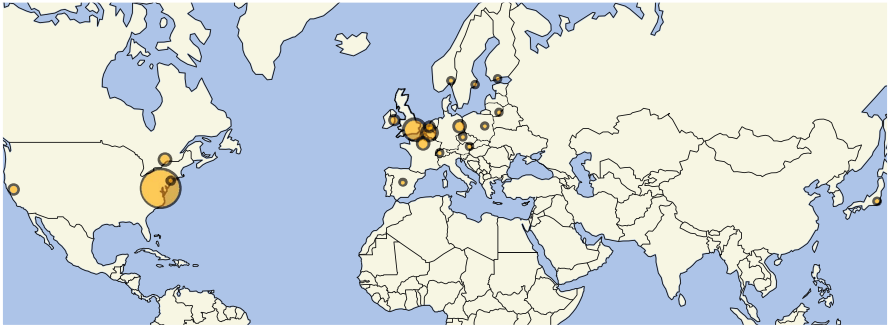

**Fig. 4** Global distribution of parliaments and legislatures from which debate transcripts are analyzed in the studies included in this review. The size of each marker is proportional to the number of studies carried out on the legislature in question

**Table 1** Sources of publicly available parliamentary and legislative debate data used in studies

| Legislature | Dataset name | Url |
|---|---|---|
| Dutch parliament | ODE | http://ode.politicalmashup.nl/data/summarise/folia |
| Norwegian parliament | Talk of Norway | http://clarino.uib.no/korpuskel/corpus-list |
| Polish Sejm | *Kronika Sejmowa* (Sejm Chronicle) | http://www.sejm.gov.pl |
| UK Parliament | Hansard | https://hansard.parliament.uk |
|  | Historic Hansard | http://hansard.millbanksystems.com |
|  | TheyWork ForYou | https://www.theyworkforyou.com/ |
| US Congress | Capitol Words | http://www.capitolwords.org/ |
|  | Congressional Record | https://www.congress.gov/congressional-record |
|  | Govtrack | https://www.govtrack.us/ |
|  | ConVote | http://www.cs.cornell.edu/home/llee/data/convote.html |
|  | SLE ConVote | http://www.cs.cornell.edu/ainur/sle-data.html |

## Research objectives, tasks, and approaches

We investigated three aspects of the studies under review: the authors stated research objectives; the task types undertaken to achieve those aims; and the approaches taken to tackling those tasks. For the latter, we report the granularity at which analysis is undertaken, the methods used, and, where applicable, the labels used to represent ground-truth sentiment/position.

### Objectives

We examined the principal stated objectives of each study in relation to the backgrounds of the researchers (see Table 2). These generally fall into two categories: (1) method evaluation, in which novel methods are presented and assessed and the

**Table 2** Included studies by background and stated research objectives

| Objective | Computer science | Political and social sciences | Multi-disciplinary |
|---|---|---|---|
| Datasets/resources | | [8, 60] | [35] |
| Linguistic analysis | [72] | | |
| Political analysis | | [8, 17, 20, 29, 53, 57, 62, 69] | |
| System/method performance evaluation | [1–7, 9, 11–16, 18, 19, 21, 30, 37, 38, 41, 48, 52, 55, 66–68, 74, 76, 78–81] | [20, 26, 32, 34, 36, 40, 47, 56, 60, 61, 69, 73] | [25, 28, 33, 51, 62, 64, 64, 65, 75] |

Individual studies may have more than one objective focus of study is the performance of the presented method or system; and (2) political analysis, in which an existing method is used as a tool to answer a political science research question, and the goal is interpretation of the chosen system's output. While the former is the focus of all CS publications and a few from political science [e.g., 11, 20], the latter was found only in papers from the social/political sciences [e.g., 17, 53]. For some political science papers, even where the primary aim appears to be the former, a common approach is to combine these objectives, first presenting a text analysis method, and then illustrating its potential by employing it to answer research questions or test hypotheses, as in Hopkins and King [26] and Proksch and Slapin [57].

Although the work of computer scientists generally focuses on system evaluation, they often state secondary application objectives which encompass motivations relating to contributions to civic technology or the development of tools for social scientists. For example, Burfoot [13] suggests that 'tools could assist researchers in understanding the nuances of contentious issues on the web by highlighting areas in which different sites or pages agree and disagree', while Budhwar et al. [12] hope that their work will 'give ordinary citizens a powerful tool to better organize and hold accountable legislators without the costs of physical presence and full-time lobbying representation'.

While the production of corpora and data sets are among the secondary contributions of many of the featured papers [e.g., 2, 66, 74], in the cases of Lapponi et al. [35] (linguistically annotated corpus) and Rauh [60] (sentiment lexicon), this is their principal objective.

Hopkins and King [26] claim that a fundamental difference between the objectives of computer scientists and political or social scientists is that, while the former are interested in making predictions about latent characteristics of individuals documents (such as sentiment polarity), the latter are more concerned with characterizing corpora or collections of documents as a whole, for example by the *proportion* of positive or negative examples it contains. This point is supported by Monroe et al. [47], who agree that individual document classification is 'inappropriate to the task', because, they suggest, under this model, representation of the whole data-generation process is backwards—where classification presumes that class labels are manifestations of the underlying latent phenomena of interest, the reality is the other way around: people first hold opinions or positions, and subsequently express them in speech or writing.

Despite this dichotomy, as we see in the next section, we do find cases of political scientists tackling classification [57] or computer scientists undertaking the scaling task from political science [21].

## Tasks

Within the overall area of sentiment analysis (as defined in Sect. 1.1), we have found that the eight types of tasks below were performed in the studies:

- Agreement and alignment detection: analysis of the similarity of the position taken by a speaker and another entity (another speaker, or a person or organisation outside of the debate in question), $n = 5$ [3, 18, 33, 48, 67].
- Emotion analysis: including emotion, anxiety, and wellbeing analysis, $n = 3$ [19, 25, 61].
- Ideology and party affiliation detection: in the literature, a speaker's party affiliation is often used as a proxy for their ideological position. This may be performed as either topic modeling or classification, $n = 14$ [6, 9, 13, 16, 17, 28, 29, 32, 35, 37, 38, 50, 51, 73].
- Opinion-topic analysis: simultaneous extraction of topics and the speakers' positions towards them, $n = 4$ [2, 50, 51, 75].
- Polarization analysis: analysis aggregated at the legislature level of the extent to which debate is polarized and its speakers are ideologically divided, $n = 2$ [29, 65].
- Position scaling: positioning of speakers or parties on a scale of one or more dimensions, $n = 11$ [8, 20, 21, 27, 34, 36, 40, 53, 56, 57, 69].
- Sentiment/opinion polarity classification: binary or ternary analysis. As votes are frequently used as opinion polarity labels, this includes the task of vote prediction, $n = 28$ [1, 4–7, 12–15, 18, 26, 30, 33, 41, 52, 55, 56, 60, 62, 64, 66–68, 72, 74, 78–81].

By far, the most frequently undertaken task is sentiment or opinion polarity classification (although this is not always named as such). In the majority of cases, this takes the form of learning from speeches the predictive features of speakers' votes [e.g., 66] or manually annotated ground-truth labels [e.g., 52]. Polarity classification is particularly prevalent in the computer science studies (24 out of 29), but despite the previously discussed claims of Hopkins and King [26] and Monroe et al. [47] that the task is incompatible with the aims of social scientists, some political scientists and multi-disciplinary teams also tackle this task [26, 56, 64].

As all the tasks undertaken concern the analysis of the positions taken by debate participants, there is considerable overlap between them. Furthermore, there is sometimes some discrepancy between the name given to a task and the actual task performed. For example, Onyimadu et al. [52] refer to the task which they perform as both 'sentiment analysis' and 'stance detection', although it could be said that they actually carry out only sentiment polarity classification as they do not specify a pre-chosen target, a requirement of stance detection (as defined by Mohammad et al. [45]). Meanwhile, Thomas et al. [74] refer to this task as 'predicting support', and Allison [5], working on the same problem and the same data set, calls it variously sentiment 'detection' and 'classification'. Although Rheault et al. [62] consider their work to be a form of emotion detection, they actually perform a form of sentiment polarity analysis at the whole legislature level, while Akhmedova et al. [4] simply refer to the task as an 'opinion mining problem'. Other terms used to refer to this include 'attitude detection' [67], 'vote prediction' [12], 'emotional polarity' measurement [62], 'predicting the polarity of a piece of text' [80], 'sentiment classification' [78, 79], and simply 'sentiment analysis' [57, 60, 64, 66, 81].

In some cases, more than one task is investigated. For example, by switching party labels for vote labels, Burfoot [13] uses the same method to perform both sentiment polarity and party affiliation (or ideology) detection. Sentiment polarity analysis is often used as part of an NLP pipeline as a sub-task of a different opinion mining task, such as agreement detection [67]. Similarly, Kauffman et al. [33] use sentiment analysis as a sub-task and the output scores as features for alignment detection, while Duthie and Budzynska [18] do similar for ethos detection, and Budhwar et al. [12] aim to predict vote outcome using the results of sentiment polarity analysis as features for the task. Conversely, Burfoot [13] applies the results of classification by party affiliation to predict speaker sentiment.

Balahur et al. [6] combine polarity with party classification, a task that we consider to be a form of ideology detection, but which they name 'source classification'. Indeed, this is another task that suffers from a lack of clarity over terminology, with some studies considering party affiliation to be a proxy for ideology [17, 29, 32, 73], while others do not make this connection, extracting information about speakers' ideologies from their sentiment towards different topics [9, 16, 51], or training a model on examples that have been explicitly labeled by ideology, and not party membership [28]. Yet others perform party classification, making no mention of the relationship between party membership and ideology [6, 13, 35, 37]. Alternatively, Abercrombie and Batista-Navarro [1] explicitly assume that membership of the same party does not guarantee homogeneity of ideologies, investigating intra-party differences of opinion and positions. Also concerned with ideology, position scaling, which we code here as a separate task, can be performed on different dimensions, one of the most common being the left- to right-wing (ideological) scale.

The literature contains several efforts to simultaneously extract topics and speakers' attitudes towards them (opinion-topic analysis). A common approach is to combine topic modeling with forms of stance detection. Nguyen et al. [50] use a supervised form of hierarchical Latent Dirichlet Allocation [10] to extract topics and polarity variables. Van der Zwaan et al. [75] generate separate topic models for different grammatical categories of words in efforts to obtain this information. And Nguyen et al. [51] perform supervised topic modeling to capture ideological perspectives on issues to produce coarse-grained speaker ideology analysis. Topic modeling has also been undertaken by Sakamoto and Takikawa [65], who use it to analyze polarization, a task also tackled by Jensen et al. [29]. Meanwhile, Vilares and He [76] also perform opinion-topic modeling to extract speakers' perspectives—'the arguments behind the person's position'—on different topics.

A number of other tasks which fit under the broader definition of sentiment analysis have also been tackled. Polarization analysis is undertaken in both Jensen et al. [29] and Sakamoto and Takikawa [65], who investigated changes in the extent to which language in Congress is polarized over time. Meanwhile, emotion detection is the subject of three studies. Dzięciątko [19] classifies speakers as expressing *happiness*, *anger*, *sadness*, *fear*, or *disgust*, while Rheult [61] attempts to identify the level of anxiety exhibited by speakers, and Honkela et al. [25] analyze a corpus of congressional speeches under the PERMA (*Positive emotion*, *Engagement*, *Relationships*, *Meaning*, and *Achievement*) model [70]. Agreement detection, an end in itself for Ahmadalinezhad and Makrehchi [3] and Kauffman et al. [33], is used by Burfoot [13], Burfoot et al. [14] and Burford et al. [15] to predict speaker sentiment, while Salah et al. [67] use agreement information to construct debate graphs. Finally, Naderi and Hirst [48] automatically compare speeches with another type of labeled text (statements from online debates) to identify positive and negative framing of arguments.

Although there exist exceptions (see above), a notable difference between the focus of tasks undertaken by NLP researchers and social scientists is that the former tend to perform analysis with regard to the target of expressed sentiment (a specific proposal [e.g., 5], piece of legislation [e.g., 74], topic [e.g., 75], or other entity), while the latter generally analyze speakers' aggregated speeches, ignoring the targets of individual contributions, and instead attempting to project actors onto a scale (such as *left–right*) [27, 34, 36, 40, 57, 69]. Grimmer and Stewart [23] note that this can be problematic as manual 'validation is needed to confirm that the intended space has been identified', and suggest automatic detection of relevant 'ideological statements' (or opinion targets) as an important challenge.

For a typology of tasks identified in this domain, see Fig. 5.

## Approaches

We consider the granularity (level of analysis), methods, features, and ground-truth labels used (in the cases of both supervised learning methods and evaluation of other methods) for each study.

### Granularity

While in other domains such as product reviews, sentiment analysis is typically carried out at the *document*, *sentence*, and *aspect* levels, here, we have found a number of approaches to segmenting the transcripts for analysis, including breaking them down to the *sub-sentence* level and aggregating sentiment over entire corpora. We also find differences in the terminology used to refer to these levels.

The vast majority of studies conduct analysis at the speech level ($n = 39$). However, 'speech' appears to mean different things in different publications, and in some, it is not immediately clear just what the unit of analysis actually is.[3] A speech may be considered to be the concatenated utterances of each individual speaker in each debate ($n = 16$). Alternatively, analysis may be conducted at the utterance or 'speech segment' level (that is, an unbroken passage of speech by the same speaker) ($n = 24$), although Akhmedova et al. [4] refer to these as 'interventions', and Bansal et al. [7] as 'individual conversational turns'. While several researchers who use Thomas et al.'s [74] *ConVote* corpus claim to analyze 'speeches', the data set (usually used unaltered) is, in fact, labeled at the speech segment level. Similar use of terminology can be found in other work, such as Vilares and He [76].

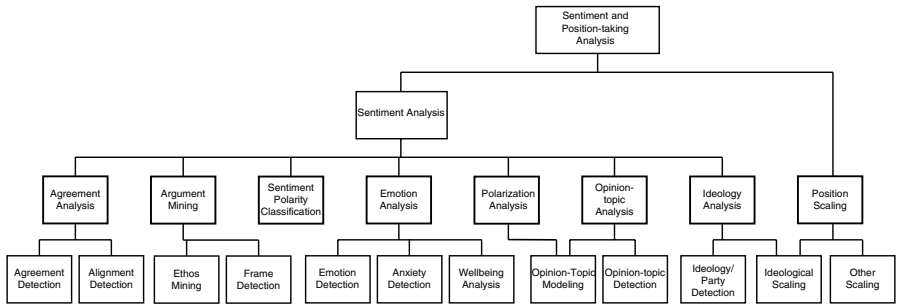

**Fig. 5** Typology of sentiment and position-taking analysis tasks performed on legislative debate transcripts, showing the eight task types identified in this review

A further eight papers report analysis at the speaker level. That is, they consider a document to be the concatenation of all speeches given by the same representative [11, 17, 33, 34, 53, 69, 73].

Other approaches are to analyze speeches at the coarser political party (or bloc/coalition) level [20, 21, 56, 57, 65, 75], or the finer sentence [18, 48, 52, 60] or phrase [29] levels. Although Rudkovsky et al. [64] detect sentiment in sentences, they aggregate these scores to provide speech-level results. Iyyer et al. [28] break speeches down to both these levels, while Naderi and Hirst [48] do so for sentences and paragraphs.

At the highest possible level of granularity, four studies consider sentiment over entire corpora. Dzieciątko [19] and Rheault et al. [62] aggregate sentiment scores for all speeches, presenting analysis of the Polish and UK parliaments respectively. Meanwhile, Honkela et al. [25] compare the overall sentiment of the European Parliament transcripts with other corpora at the whole data set level, and, in addition to party-level analysis, Sakamoto and Takikawa [65] compare the polarity of Japanese and US datasets.

Finally, Ahmadalinezhad and Makrehchi [3] consider each document to be a 'conversation between two individuals'—that is, the two parties' combined utterances—to classify these as being either in agreement or disagreement.

Overall, computer scientists tend to work at finer-grained levels (speech, speech segment, paragraph, sentence, or phrase), while in political science, the preferred units of analysis are the actor (or individual politician, whose contributions are pooled together), which is the target of most work on position scaling, a task very much associated with that field. This confirms to some extent the assertion of Hopkins and King [26] that, while computer scientists are 'interested in finding the needle in the haystack, ... social scientists are more commonly interested in characterizing the haystack'. Exceptions, from the political and social sciences, are Iliev et al. [27], and Hopkins and King [26]—who actually propose a method of optimizing speech-level classification for social science goals, and from computer science, Glavaš et al. [21], who also tackle the position scaling problem.

---

[3] In several cases, it has been necessary to contact the authors for clarification or to manually examine the data sets used to obtain this information.

## Methods

A wide range of approaches are used, but these can be grouped into the following five main methods (of which some publications use more than one):

- Dictionary-based: using lexicons to assign sentiment scores, $n = 16$ [3, 6, 12, 16, 18, 19, 25, 52, 53, 56, 60, 62, 66–68, 76].
- Statistical machine learning: used to learn a predictive function based on the data, $n = 45$ [1–7, 9, 11–18, 21, 25–28, 30, 32, 33, 35, 37, 38, 41, 48, 50, 51, 55, 61, 64, 65, 67, 72–76, 78–81].
- Rule-based systems: including hand-crafted heuristics and rule-based machine learning classification, $n = 3$ [4, 52, 66].
- Similarity comparison: use of measurements such as cosine similarity/difference to link or group examples, $n = 9$ [3, 6, 13, 15, 21, 37, 48, 61, 67].
- Word frequency analysis methods: simple counting of $n$-grams used to position examples on a scale, $n = 10$ [8, 20, 29, 34, 36, 37, 40, 56, 57, 69].

Overall, the scaling task is approached using methods based on simple unigram counts (although a few make use of machine learning approaches [e.g., 17, 26, 61]). The scaling task comes in two varieties: supervised, in which target speeches ('virgin texts') are compared to reference texts, and unsupervised, such as the Wordfish package introduced by Proksch and Slapin [57], and used by Schwarz et al. [69]. Glavaš et al. [21], in the only study conducted from an NLP perspective that takes on the position scaling problem so favored by political scientists, use a combination of semantic similarity measurement and harmonic function label propagation, a semi-supervised graph-based machine learning algorithm.

In total, roughly three quarters of included studies ($n = 45$) make some use of machine learning, and within this area, there are a multitude of different approaches (see Fig. 6). These can be broadly categorized as supervised learning ($n = 40$), semi-supervised ($n = 1$), or unsupervised ($n = 11$) methods. Supervised methods are the preferred approach for text classification, and a wide variety of algorithms are used, including logistic regression ($n = 10$), naive Bayes ($n = 6$), decision trees ($n = 3$), nearest neigbor ($n = 3$), boosting ($n = 1$), a fuzzy rule-based classifier ($n = 1$), maximum entropy ($n = 1$), nearest class classification ($n = 1$). A further 11 studies make use of neural networks, which range in complexity from 'vanilla' feed-forward networks such as the multi-layer perceptron to convolutional and recurrent neural networks, including the use of long short-term memory units (LSTMs) (see Fig. 5). Of these, six are concerned with sentiment polarity analysis and four with ideology detection, which, as previously discussed, are highly similar classification tasks, performed with different class labels. The exception is Duthie and Budzynska [18], who use recurrent neural networks modules in their ethos mining task.

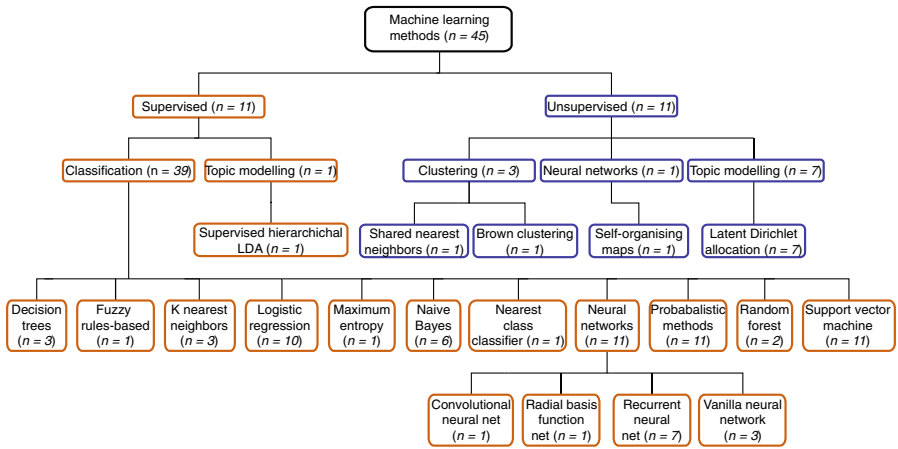

**Fig. 6** Machine learning methods used in the included publications for sentiment and position analysis

Rather than testing different classification algorithms, some work focuses on the use of different regularization methods (for logistic regression) [79–81], while other approaches to improving performance of classifiers include Boosting [12], and Collective Classification [14].

Much of the work on unsupervised learning focuses on use of topic modeling methods, the majority of which are variations of LDA. The cross-perspective topic model of van der Zwaan et al. [75], for example, generates two topic models from the data: one over nouns to derive topics, and the other over adjectives, verbs, and adverbs, which is intended to produce opinion information.

Additionally, seven studies model debates as networks of connected speakers and employ graph-based methods. Bansal et al. [7], Burfoot [13], Burfoot et al. [14, 15], and Thomas et al. [74] all approach sentiment polarity classification as a *minimum cuts* graph partition problem. Chen et al. [16] construct an 'opinion-aware knowledge graph', propagating known ideology information through the graph to infer opinions held by actor nodes towards entity nodes. For position scaling, Glavaš et al. [21] use similarity measurements as edges between documents, and propagate the scores of known 'pivot texts' to the other nodes.

## Language models and feature selection

Although analysis of debate transcripts necessarily utilizes textual features derived from the speeches, there are a variety of approaches to how this is modeled, and which types of features are selected. In terms of language modeling, the majority of studies represent the text as bags-of-words.[4] However, some add contextual information with use of word embeddings. While their use is generally restricted to studies from computer science, Rudkovsky et al. [64] (multi-disciplinary) and Glavaš et al. [21] (computer science) explore the use of word embeddings for aims normally associated with the social sciences. Relatively little linguistic or structural analysis is undertaken. Exceptions are Ji and Smith [30], who find that the application of Rhetorical Structure Theory does not produce satisfying results in this domain, because its 'discourse structure diverges from that of news', and Balahur et al. [6], who extract parse trees to determine the target of speech sentiment.

In addition to textual information, many studies also make use of metadata features of the speakers, as well as other features such as those derived from the structure of the debates. The following six categories of features are used:

- Textual: including word (all studies) or character [32] *n*-grams, custom dictionary keyword features [12], words from particular grammatical categories [28, 32, 35, 47, 48, 52, 72, 75], presence of questions [12], word embeddings [9, 21, 28, 30, 38, 48, 61, 62], sentence [64] embeddings, and parse trees [6, 28, 30].
- Debate discourse features: including citations [13, 14, 37], interruptions and speech length [12], *n*-grams from other neighboring sentences [78], and utterance statistics (number and duration of speaker's utterances) [33].
- Speaker metadata features: including bill authorship [12], debate and speaker IDs [1, 66], debate type [35], donations [11, 33], geographic provenance [35], party affiliation [1, 2, 13, 33, 64–66], and gender [35].
- Polarity scores: the output of opinion polarity analysis used as a feature for prediction of another phenomena (such as sentiment polarity [9], voting intention [12], argument frame [48], or alignment with another entity [33]).
- Relational and knowledge graph features: features based on the relationships between different speeches or speakers, speakers and other entities (such as the targets of expressed opinion), or the measured similarity between speeches [13–16, 74].
- Speaker vote: speaker's votes *for/against* a motion or piece of legislation, used as a feature to identify their ideology [34, 69].

A key decision for researchers is that of whether or not to include non-textual, metadata features, and the answer to this is usually driven by their research objectives. In some studies, particularly those from political science focused on position scaling, the object may be to examine intra-party differences or to compare speech and vote behavior, in which cases features such as party affiliation or vote are the dependent variable under observation, and cannot be used as features for analysis. For classification, while some researchers compare performance with and without this additional information [e.g., 1, 2, 67], others prefer to exclude them entirely to make their methods more generalizable to debates from other domains such as online debates, which do not have access to such information [13, 74].

---

[4] That is, unstructured, unordered arrays of *n*-gram (term) counts.

**Ground-truth labels**

Depending on the nature of the task being tackled, for supervised classification methods, and in some cases, validation of unsupervised methods, several different data sources are used to represent the ground-truth sentiment or position. In most cases, researchers opt to make use of some pre-existing data, with the most common being the speakers' roll-call or division[5] votes ($n = 21$). Here, the most common approach is to consider each speaker's vote on a given debate to represent the ground-truth position taken in their speeches, which may be analyzed at the whole (concatenated) speech level (as in Salah [66]) or broken down into smaller units (as in Thomas et al. [74]), whereby each vote label is attached to multiple examples. A general difference in approach is that, while computer science studies use these as ground truth, political scientists tends to view speech and vote as unconnected, and even explicitly compare the two on this basis, as in Schwarz et al. [69]. Lowe and Benoit [40] use human annotations for validation of the output of their scaling method. Whether or not votes are actually reliable as ground-truth is a matter of contention. Although some computer science studies assume this to be the case [e.g., 66, 68, 74], Schwarz et al. [69] compare speeches to votes for scaling, producing quite different results, and Abercrombie and Batista-Navarro [1] who compare votes with human produced labels, conclude that it is the latter which more closely reflect sentiments expressed by speakers.

An alternative approach is to use manually annotated labels. While some researchers make use of already existing expert annotations, such as the Chapel Hill expert surveys[6] [21, 57], others produce labeled data sets specifically for their purposes. Onyimadu et al. [52] and Rauh [60] both had in-house annotators label speech sentences as being *positive*, *negative*, or *neutral*, while Rheault [61] used crowd-sourced coders to label sentences as *anxious* or *non-anxious*. Rudkovsky et al. [64] also use crowd-sourced labels, but for evaluation rather than training purposes, in their case to assess negativity detection. To create labels for the validation of their scaling of speakers' positions towards a given topic, Frid-Nielsen [20] had experts follow the coding scheme of the Comparative Manifestos Project[7] to produce policy position labels, although the reliability of these is also controversial [44].

Other data used as ground-truth labels are the speakers' DW-nominate scores (scores derived from congressional legislators' voting records[8]) [17, 51], their constituency vote shares [73], 'issue' labels from the Library of Congress's Congressional Research Service[9] [11], word perplexity [75], sentiment analysis scores obtained from prior experiments on the same data [72], and party affiliations ($n = 13$). While the latter are widely used as a proxy for speaker ideology [9, 29, 38], Hirst et al. [24] suggest that party membership is actually a confounding factor for this task.

---

[5] Terms used in the US Congress and UK Parliament, respectively.
[6] https://www.chesdata.eu/.
[7] https://manifesto-project.wzb.eu/.
[8] https://legacy.voteview.com/dwnomin.htm.
[9] https://www.loc.gov/crsinfo/.

## Performance and outcomes

With the research reviewed here having such varied objectives and undertaking many different analysis tasks, it is not possible to directly compare the reported performances of the methods proposed. Nevertheless, in this section, we attempt to summarize some conclusions of the included studies that are potentially relevant to future work in this area.

For classification, machine learning methods, and particularly neural networks, seem to outperform other approaches. Here, just as in other domains, such as product reviews, dictionary-based sentiment analysis methods appear to have been superceded by machine learning approaches. In a direct comparison, Salah [66] found that machine learning classification methods outperform those utilizing both generic and parliament-specific lexica, while Balahur et al. [6] improved lexicon-based performance with the addition of a support vector machine classifier. Given this, and also considering the conclusion of Allison [5] that 'classifier choice plays at least as important a role as feature choice', which learning algorithms should be selected for classification in this domain? In the work reviewed here, support vector machines, used in 29 of the studies, are the most popular option—both as a common baseline, and as a default algorithm choice. Although, in NLP in general, the last decade has seen an explosion in interest in deep learning methods, here, we see relatively little use of neural network-based machine learning. Those studies that do directly compare the performance of such methods with other classifiers suggest a tendency towards better performance using neural networks [1, 12, 28, 38].

For position scaling, political and social scientists do not tend to place the same emphasis on performance metrics such as accuracy, preferring to make comparisons between output manual analyses to investigate theory-based hypotheses. Indeed, discussion of technical performance in these papers often focuses on whether or not computational text analysis is valid at all when compared with expert examination. In this respect, [17, 20, 36] conclude that, with some caveats, it is a legitimate approach (Lowe and Benoit [40] note that their method appears to position some speakers on a different dimension to that of their expert analysis). In the one computer science paper to tackle this problem, Glavaš et al. [21] report equally promising results on mono- and multilingual data, as well as superior performance using word embeddings over a bag-of-words model.

In the reviewed publications, a large range of feature types are extracted from the transcripts. Most studies rely primarily on the bag-of-words model, and for textual features, the benefits of adding higher order $n$-grams (bi-, tri-grams, etc.) appear inconclusive. While Plantié et al. [55] report improved performance with the addition of bigrams to their feature set, Abercrombie and Batista-Navarro [1] do not see significant improvement with the use of bi- and tri-grams. With the most common method of $n$-gram feature selection being TF-IDF weighting, Martineau et al. [41], noting that IDF favors rare features, find that, for the relatively homogenous domain of a particular parliament's transcripts, their alternative Delta TF-IDF representation leads to better classification performance.

As we have seen, the appropriateness of using metadata features depends on the objectives of the research. However, if optimal classification performance is the goal and information regarding the speakers' party affiliations is available, this has been found to be highly predictive of expressed sentiment [1, 66]. Inter-document relationship information regarding agreement between speakers also assists in sentiment polarity classification, and has been applied successfully by Bansal et al. [7], Burfoot [13], and Thomas et al. [74], as has network information [14, 15]. The latter show that it is possible to model these relationships for any data set using *n*-gram overlap. In another approach to modeling debate structure, Balahur et al. [6] use dependence parsing to find targets, which seems to improve classification and helps to balance results obtained in the positive and negative classes. While Iyyer et al. [28] also report success in using parse trees as features for classification with a recurring neural network, Ji and Smith [30] do not find improvement in the parliamentary domain (although they do in news articles).

When it comes to representing ground-truth, votes are not necessarily indicative of the opinions expressed in speeches, but for speech-level polarity analysis, they can be a convenient option. The results of computational analysis by Schwarz et al. [69] supports manual analysis in political science [58] to indicate that representatives position themselves differently in their speeches than in their voting behavior. However, the relatively small difference between votes and manual annotations (less than four per cent of their corpus) found by Abercrombie et al. [1] suggests that relatively small gains are to be had by investing in human labeling where other forms of class label are available.

A number of observations arise about the use of language in this domain. For the UK Parliament, Onyimadu et al. [52] find that 'compound opinions', sarcasm, and comparative structures are all confounding elements for classifiers. In German, Rauh [60] notes that 'positive language is easier to detect than negative language', while Salah et al. [67, 68] make a similar observation for the UK Hansard transcripts. The latter study explains this phenomena as an artifact of the 'polite parliamentary jargon' used in Parliament. This point is also backed up by Abercrombie and Batista-Navarro [1], who observe that the most indicative features, even of negative polarity, are words not typically thought of as conveying negativity. Where negative adjectives and verbs *are* present, Sokolova and Lapalme [72] find that these are highly discriminative features.

## Conclusion and discussion

### Scope for further inter-disciplinary collaboration

Considering the nature of the problem at hand—computational methods for the analysis of political text—it is somewhat surprising how little crossover can be found in this domain between ideas from computer science and political science, and how seldom the methods used by researchers from these different fields are adopted by researchers from the other disciplines. As an explanation for this, Hopkins and King [26], Monroe et al. [47], and Lowe and Benoit [40] provide insights into the differing aims of the two fields. However, despite these differences, researchers in computer science may well be able to benefit from the theoretical expertise of political and social scientists, such as the rigourous labeling schema and expertly coded corpora already existing in the field. Similarly, more political and social scientists could consider going beyond the simple bag-of-words *n*-gram language models which they currently rely on to investigate the use of more advanced NLP methods of representing text and handling feature sparsity in natural language, such as word embeddings.

## Standardization of terminology

A problematic issue that arises from surveying the work included in this review is the wildly inconsistent use of terminology, even within each of the research fields represented here. There is a clear need for greater agreement on which terms to use to refer to the affective targets of interest and the names of the tasks designed to analyze them, as well as the varying levels of granularity at which analysis is performed. These inconsistencies often mean that it is difficult—or even, without further investigation, impossible—for the reader to understand just what is done in a given study.

## More fine-grained level of analysis

Studies included in this review approach analysis of legislative transcripts at a wide variety of granularities, from the phrase level to comparisons aggregating sentiment over entire corpora. However, for the sake of convenience, and to make use of existing labels such as votes, the majority conduct analysis at the speech level, or even if they do so at a more fine-grained sentence or phrase level, they tend not to consider the discourse structure of the debates. As Burfoot [13] points out that parliamentary and legislative debates are complex, with many topics discussed and sentiment directed towards varying targets in ways that a document level classifier can struggle to identify. There is therefore room to develop more complex analyses, capable of recognizing the relationships between entities and targets in fine-grained sections of the transcripts, perhaps using argument mining methods that harnesses theories from fields such as communication theory [e.g., 48] or even philosophy [e.g., 18] to explore the relationships between actors, opinion, targets, and other entities in debates.

There have also been few attempts to link expressed opinion with topic information. While there are some efforts to do so from debate motions [2], at the political party level [75], and as a form of perspective analysis [76], as well as by scaling on pre-defined topic dimensions [53], the majority of studies simply conduct analysis of sentiment towards a target, such as a Bill or motion, the topic of which is unknown. To provide truly useful information, it may make sense to focus efforts on the extraction of topic-centric opinions and to conduct analysis at the level at which different topics are found in the data.

## Use of ground-truth labels

While the majority of studies that focus on supervised classification rely on votes as ground-truth labels, it is debatable whether these actually represent the target phenomena—the opinion or position taken by the speaker. Manual analysis in political science [58] certainly suggests that, in many legislatures, representatives express different positions in speech than in their votes, a point supported by Schwarz et al. [69], who compare the scaling of speeches and votes. However, as Abercrombie and Batista-Navarro [1] point out, gains made from seeking more reliable ground-truth may be small and not worth the associated costs in a practical setting. An alternative approach, for which there is still plenty of scope for further research, is to develop semi-supervised or unsupervised approaches, which require few or no labels.

## Other possibilities for the direction of future work

There also exist other possible directions for future work. With an increasing quantity of transcript data becoming available, in the case of some parliaments stretching back over hundreds of years, one such possibility is the analysis of language change over time. While Diermeier et al. [17] suggest using changes in classification performance to infer changes in agenda control, and Kapočiūtė-Dzikienė and Krupavičius [32] found that performance worsened when transcripts from different sessions were used for training and testing, language drift in parliamentary debates remains relatively unexplored. Although Ahmadalinezhad and Makrehchi [3] note a performance drop when training and testing on different debate data (Canadian Hansard and US election debates), domain adaptation and inter-legislature transfer learning also remains under-explored. Additionally, given the successes achieved with neural networks and deep learning in other domains, as well as the results reported by studies that use such methods, there would appear to be considerable scope for further investigation of their application to legislative debates. Finally, while some of the included studies mention potential applications of their work for civic technology [e.g., 12, 13], with the exception of Bonica's [11] CrowdPac Voter Guide, as far as we know, the methods used are not currently being applied to any real-world systems. There is, therefore, room to explore the application of the approaches used to the area of civic technology, and provide tools that could genuinely assist people in processing information about their elected representatives.

The computational analysis of sentiment and position-taking in parliamentary debate transcripts is an area of growing interest. While the researchers working on this problem have varied backgrounds and objectives, in this review, we have identified some of the common challenges which they face. With the majority of work emanating from computer science focusing on unknown targets (Bills or debate motions, the topic of which is not assessed), and political scaling being conducted on very coarse-grained scales (*left-right*, *pro/anti-EU*), there has thus far been little effort to direct efforts towards examining the targets of the opinions expressed. For the aims of both political scholarship and civic technology, what is required in many cases is identification of these targets, namely the policies and policy preferences that are discussed in the legislative chambers. It is our belief, therefore, that future work should be directed towards such target-specific analyses.

## Compliance with ethical standards

**Conflict of interest** The authors declare that they have no conflict of interest.